%% file: root.tex
\documentclass[letterpaper, 10 pt, journal, twoside]{IEEEtran} % Use this command for final RAL version
\usepackage{graphics} % for pdf, bitmapped graphics files
\usepackage{epsfig} % for postscript graphics files
\usepackage{amssymb}  % assumes amsmath package installed

\usepackage{algorithm,algorithmic}  % Added by Jie
\usepackage[svgnames,table]{xcolor} % to set the color of tables

\definecolor{rgb1}{RGB}{214,  38, 40}   % 1 ceiling
\definecolor{rgb2}{RGB}{43, 160, 4}     % 2 floor
\definecolor{rgb3}{RGB}{158, 216, 229}  % 3 wall
\definecolor{rgb4}{RGB}{114, 158, 206}  % 4 window
\definecolor{rgb5}{RGB}{204, 204, 91}   % 5 chair
\definecolor{rgb6}{RGB}{255, 186, 119}  % 6 bed
\definecolor{rgb7}{RGB}{147, 102, 188}  % 7 sofa
\definecolor{rgb8}{RGB}{30, 119, 181}   % 8 table
\definecolor{rgb9}{RGB}{160, 188, 33}   % 9 tvs mod
\definecolor{rgb10}{RGB}{255, 127, 12}  % 10 furn
\definecolor{rgb11}{RGB}{196, 175, 214} % 11 objects

\newcommand{\etal}{\textit{et al}.}
\newcommand{\ie}{\textit{i}.\textit{e}.}

\author{Yu Liu$^{1,3^*}$, Jie Li$^{2^*}$,
Xia Yuan$^{2}$, Chunxia Zhao$^{2}$, Roland Siegwart$^{3}$, Ian Reid$^{1}$ and Cesar Cadena$^{3}$
\thanks{
\textit{Yu Liu and Jie Li contributed equally to this work, part of the work was done when Yu Liu doing internship at ETH ASL. (Corresponding author: Cesar Cadena.)}
}
\thanks{$^{1}$Y. Liu and I. Reid 
are with School of Computer Science, The University of Adelaide,
5005, North Terrace, SA 
{\tt\small yu.liu04@adelaide.edu.au}}%

\thanks{$^{2}$J. Li, X. Yuan and C. Zhao 
are with School of Computer Science and Engineering, 
Nanjing University of Science and Technology, Nanjing, 210094, China 
{\tt\small jieli\_cn@163.com}}%

\thanks{$^{3}$R. Siegwart and C. Cadena 
are with Autonomous Systems Lab, ETH Zurich, 
Leonhardstrasse 21, 8092, Zurich 
{\tt\small cesarc@ethz.ch}}%
} %Use only for final RAL version. 

\title{\LARGE \bf
Depth Based Semantic Scene Completion with Position Importance Aware Loss
}  %Use for final RAL version

\begin{document}

\maketitle
% \thispagestyle{empty}
% \pagestyle{empty}

%%%%%%%%%%%%%%%%%%%%%%%%%%%%%%%%%%%%%%%%%%%%%%%%%%%%%%%%%%%%%%%%%%%%%%%%%%%%%%%%
\begin{abstract} Semantic Scene Completion (SSC) refers to the task of inferring the 3D semantic segmentation of a scene while simultaneously completing the 3D shapes. 
We propose PALNet, a novel hybrid network for SSC based on single depth.
PALNet utilizes a two-stream network to extract both 2D and 3D features from multi-stages using fine-grained depth information to efficiently captures the context, as well as the geometric cues of the scene.
Current methods for SSC treat all parts of the scene equally causing unnecessary attention to the interior of objects.
To address this problem, we propose \textit{Position Aware Loss(PA-Loss)} which is position importance aware while training the network. 
Specifically, PA-Loss considers Local Geometric Anisotropy to determine the importance of different positions within the scene. 
It is beneficial for recovering key details like the boundaries of objects and the corners of the scene. 
Comprehensive experiments on two benchmark datasets demonstrate the effectiveness of the proposed method and its superior performance. 
Code and demo\footnote{Video demo can be found here: \textcolor{cyan}{
https://youtu.be/j-LAMcMh0yg}} are avaliable at: https://github.com/UniLauX/PALNet.
\end{abstract}

% Keywords appear just beneath the abstract. Use only for final RAL version.   

%%%%%%%%%%%%%%%%%%%%%%%%%%%%%%%%%%%%%%%%%%%%%%%%%%%%%%%%%%%%%%%%%%%%%%%%%%%%%%%%
% Drop letter for first word of the Introduction  % Here we have the typical use of a "T" for an initial drop letter 
% and "HIS" in caps to complete the first word. \IEEEPARstart{T}{his} is the first sentence of my Introduction. % Use only for final RAL version

\section{INTRODUCTION}
% p1.1 SSC and its board applications
\IEEEPARstart{S}{emantic} scene completion (SSC)~\cite{choi2015indoor, Gupta2015, song2017_SSCNet} is composed of scene completion and semantic labeling, which targets at inferring the completed 3D shape and the semantic categories of the occupied voxels (grid units) in the scene simultaneously. 
As illustrated in Fig.~\ref{fig:teaser}, given a single-view depth, SSC assigns a class label to every voxel including the surface of the scene and the occluded parts in the frustum.
SSC is useful and widely applied in numerous real-world applications such as virtual reality~\cite{chen2017semantic, van2010survey},
robot grasping~\cite{varley2017shape},
automatic navigation~\cite{vineet2015incremental},
\textit{etc}.

\input{figs/teaser.tex}

3D convolutional neural network (3D-CNN) is frequently used in the task of 3D scene prediction. Since 3D-CNN requires a regular grid as input, voxels are naturally chosen to represent the 3D scene in these 3D-CNN based methods, and different loss functions are employed for training them~\cite{song2017_SSCNet,guo2018_VVNet,Tchapmi2017_SEGCloud}.

% 1.2 Motivation for PA-loss
A limitation of previous methods is that they ignore the position importance of the voxels. 
Usually, all of the voxels are treated equally regardless of their positions, appearance, and shape. 
However, from the human perception system~\cite{lindsay2013human, desingh2013depth}, when we recognize objects, the two most prominent features that make the target object distinguish from other surrounding objects are the corners and uneven surfaces. 
Furthermore, there are a large amount of voxels located within the objects which are likely to share the similar texture and are redundant for classification task. 
A relatively small portion of the voxels (at the edges of a scene or the corners of the objects) are the critical clues for identifying the classes.
Based on these observations, we propose Local Geometric Anisotropy (LGA) to quantitatively determine the importance of different positions, and further introduce \textit{Position Aware Loss (PA-Loss)} which gives different importance to voxels based on their LGA.
With PA-Loss, the network is forced to pay more attention to the surfaces, edges, and corners which leads to slightly faster convergence and a boost in performance of the SSC task.

% p2.2 3D convolution are memory consming, which limit the resolution of 3D grid.
Another limitation of the previous methods is that they do not use the complete depth information.
% Moreover, the voxelization based methods cannot make full use of the depth information.
During voxelization, several pixels in depth can be projected into one voxel in the 3D grid volume and due to the memory limitation, the input resolution used in 3D-CNN is unlikely to be very large, which leads to a coarse voxel level representation.
In particular, a common voxelized input for 3D-CNN models is $240\times 144\times 240$ ~\cite{song2017_SSCNet,Garbade2018_twoStream,zhang2018efficient}, while the original input depth has resolution of 640 $\times$ 480 or much higher. 
In this paper, we take both the advantages of the full resolution depth and the encoded voxels by designing a hybrid network structure with two input branches.
In the other branch, we feed the network directly with the original depth to avoid the loss of detailed information.
In another branch, we adopt the 3D voxels as the input to take advantage of the Truncated Signed Distance Field (TSDF) encoding~\cite{izadi2011kinectfusion}. Through the use of differentiable 2D-3D projection layer, the 2D CNN and 3D CNN blocks are effectively combined within the network.

% 2.3 Contributions
The main contributions of this paper are two-fold.

\begin{itemize}
\item Firstly, we propose a novel PA-Loss for SSC task, which emphasizing the rare voxels that are located on the surface or corners of the scene, while diluting the voxels which carry redundant information within the objects. Our experiments indicate that PA-Loss has the benefit of slightly faster convergence for training and can achieve better performance than previous works.

\item Secondly, based on single-view depth, we propose a hybrid network, which takes full advantage of both fine-grained depth and TSDF. The detail information extracted from the full resolution depth is beneficial for semantic labeling as well as scene completion, which also distinguish our approach from the existing mainstream approaches that only use the voxelized TSDF as the sole input.

\end{itemize}

% Experimental results on NYU~\cite{silberman2012NYUdataset} and NYUCAD~\cite{firman2016NYUCAD} datasets demonstrate that our method achieves state-of-the-art performance.

% The rest of this paper is organized as follows. Section~\ref{relatedworks} briefly summaries the related works, Section~\ref{methodology} introduces the methodology. Section~\ref{experiments} presents the experimental results on NYU and NYUCAD datasets.
% Section~\ref{ablationstudy} gives abalation study of different parts.
% Section~\ref{conclusion} summarizes our findings and concludes with future research interests. 

%%%%%%%%%%%%%%%%%%%%%%%%%%%%%%%%%%%%%%%%%%%%%%%%%%%%%%%%%%%%%%%%%%%%%%%%%%%%%%%%
\section{Related work}
\label{relatedworks}
\subsection{Semantic Scene Completion}
% \vspace{-1mm}
% Deep learning based methods
Recently several methods have been proposed for SSC using deep learning techniques~\cite{song2017_SSCNet,Garbade2018_twoStream,zhang2018efficient,guo2018_VVNet}.
Among them, the most representative work is the SSCNet~\cite{song2017_SSCNet} which conducts the semantic labeling and scene completion simultaneously and also proves that these two tasks can benefit from each other. 
SSCNet takes advantage of TSDF and uses voxels to represent the 3D space. 
Although better results have been achieved compared with the previous methods, SSCNet ignores the fine-grained information of depth.
% , so that the final precision still has an ample space to improve.
Zhang~\etal~\cite{zhang2018efficient} introduces spatial group convolution (SGC) to reduce the computation costs but with poor performance than SSCNet~\cite{song2017_SSCNet}. SEGCloud~\cite{Tchapmi2017_SEGCloud} employs fine-grained 3D point as input but the computing and memory costs are incredibly high. % --- VVNet
VVNet~\cite{guo2018_VVNet} combines 2D-CNN and 3D-CNN by replacing some 3D volume layers with the corresponding 2D view network layers which leads to a much accurate and efficient network compared with SSCNet. However, it discards TSDF, which can provide the prior knowledge about the space encoding and is vital to distinguish between the empty and occupied parts of the scene.

Two representives of generative models for SSC are 3D-GAN~\cite{3dgan} and ASSC~\cite{wang2018adversarial}.
3D-GAN~\cite{3dgan}
uses volumetric convolutional networks to generate 3D objects from a probabilistic space.
% , and is limited to generate high-quality 3D objects.
% --->
% It can only generate objects and cannot generate complete scenes.
% 
ASSC~\cite{wang2018adversarial} applies auto-encoder to encode the latent context of the single-view depth and uses a 3D generator
% for self-reconstruction. 
to rebuild the 3D complete scene.
% However, it will result in the loss of location information due to the use of the fully connected layers.
% --->
% The location information is not directly passed to the 3D generator, but can only be passed indirectly through the hidden layer.

% P2.2 RGB+D for SSC: TS3D, DDR-SSC
To enrich the input information and boost the accuracy of SSC, TS3D~\cite{Garbade2018_twoStream} and
DDR-SSC~\cite{li2019rgbd} proposed to add a RGB branch in addition to the voxel branch, which introduce extra network or parameters, and are less accurate than our method. 

Different from the previous methods, our proposed \textit{Position Aware Loss Network (PALNet)} takes advantages of both the fine grained depth and the TSDF encoded 3D volume.
Specifically, we formulate the depth detail as a vital ingredient in the proposed network and make full use of the high-resolution depth by a 2D CNN.
In the 2D CNN, a differentiable projection layer is employed to accurately project features in 2D space to the corresponding locations in 3D volume. 
Moreover, the encoded TSDF provides the prior geometric knowledge of scene which contributes to a much accurate model for SSC.

%%%%%%%%%%%%%%%%%%%%%%%%%%%%%%%%%%%%%%%%%%%%%%%%%%%%%%%%%%%%%%%%%%%%%%%%%%%%%%%%
\input{figs/VoxelPosition.tex}

\input{figs/hist_percentage.tex} 

\subsection{Loss Function for 3D Dense Prediction}
Compared to 2D image segmentation, 3D dense prediction has three characteristics.
Firstly, the number of voxels in 3D dense prediction is much larger than the amount of pixels in the 2D image segmentation.
Secondly, the number of voxels ranges a lot among objects with different sizes.
Finally, the number of voxels
outside the objects is far beyond that of inside the objects.
And we further observed that in 3D semantic scene completion, voxels at different positions make different contributions as well as deliver various training difficulties to the scene understanding as mentioned before. Based on the above observations, it is thus vital to choose a suitable loss function to train the 3D network effectively with the consideration of voxel-wise data-balance. There are many classic loss functions available to train 3D networks.

\subsubsection{Cross-entropy Loss}
Extended from 2D vision tasks, cross-entropy loss can be used in 3D dense prediction. In essence, it treats all the predicted targets equally and is proved less efficient in 3D tasks~\cite{milletari2016vnet, song2017_SSCNet}.
 
\subsubsection{Weighted Cross-entropy Loss}
Weighting factor $w_c\in [0, 1]$ is introduced based on cross-entropy loss to handle the imbalance problem. 
Weighted cross-entropy loss can emphasize the importance of classes with rare samples, while it relies on manually set weight parameters.
A compromise approach instead of manual selection of weights is to set $w_c$ as the inverse frequency for the corresponding class. 
And it can only handle category-level data imbalance, but not voxel-wise imbalance.
% and is limit to handle category-level data imbalance.

\subsubsection{Focal Loss}
Focal loss aims to address the data imbalance in object detection especially when the dataset contains too many easy negatives that contribute no meaningful learning signals. 
However, one limitation of using focal loss is that it underestimates the importance of well classified samples~\cite{nguyen2018u,redmon2018yolov3}. Also, the training process becomes sensitive to incorrectly labeled samples~\cite{guo2018locally}.

\subsubsection{Dice Loss}
Dice loss~\cite{milletari2016vnet} is proposed to address the data imbalance in volumetric medical image segmentation. 
It is a good choice to address the imbalance between the foreground and background in binary segmentation, but does not generalize well to multi-category segmentation. 
Besides, it is not as easy to optimize as cross-entropy loss, as its gradient may blow up to some enormous value when both the value of the target label and the prediction are small.

In summary, none of those loss functions can take into account the importance of different positions. 
In this paper, we propose Local Geometric Anisotropy (LGA) to determine the importance of geometric information contained in different voxels, and LGA is then used to calculate the weight factor for cross-entropy loss to form the proposed PA-Loss, which fully considers the impact of each element, leads to 
% fast training convergence and 
the better performance compared to other loss functions.

\input{figs/NetworkStructure.tex}
%%%%%%%%%%%%%%%%%%%%%%%%%%%%%%%%%%%%%%%%%%%%%%%%%%%%%%%%%%%%%%%%%%%%%%%%%%%%%%%%
\section{Methodology}
\label{methodology}
We propose a framework for semantic scene completion in which: (1) Voxel-wise data imbalance is handled by the proposed PA-Loss. (2) Both depth details as well as geometry prior can be fully utilized by the novel PALNet as illustrated in Fig.~\ref{fig:NetworkStructure}. The details of the PA-Loss and PALNet are presented in the following subsections.

\subsection{Position Importance Aware Constraint}

\subsubsection{ Local Geometric Anisotropy (LGA)}
The geometric information contained in voxels at different positions has high variability.
In particular, the voxels inside the same object are more homogeneous and likely belong to the same semantic category as their surrounding voxels.
Meanwhile, the voxels at the surfaces, edges, and vertices have richer geometric information
(have different semantic labels with their surroundings)
than those that are inside the objects. 
We refer to the semantic difference between the current voxel and its surrounding neighbors as Local Geometric Anisotropy.

% To measure the LGA of a voxel with a specific semantic label, all other voxels which have different semantic labels are considered as another category. 
% In this way, the LGA defining can be viewed as a  binary category problem. 
To measure the LGA of a voxel with a specific semantic label, we focus on voxels that belong to the current semantic category and treat all other voxels with different semantic labels as another category.
Specifically, the voxels which are sharing the same category with the current one are denoted `occupied' voxels and the voxels that belong to other categories are denoted `unoccupied' voxels, as shown in Fig~\ref{fig:VoxelsPosition}.
Given a voxel $p$, its LGA is calculated based on the 6-neighbor voxels, which is expressed as Eq~\ref{Eq:lga}:

% we also enumerate all possible cases (row1) and show an example of the LGA of each case.

\begin{equation}\label{Eq:lga}
M_{ LGA }=\sum _{ i=1 }^{ K }{ \left( { c }_{ p }\oplus { c }_{ q_{ i } } \right)  } 
\end{equation}
where $c_p$ is the semantic label of current voxel $p$, ${ q }_{ i }$ is one of its $K$ neighbours and $i\in \{ 1,2,\cdots ,K\}$,
${ c }_{ q_i }$ is the semantic label of ${ q }_{ i }$. And $\oplus$ is the exclusive or (XOR) operation.
If voxel $p$ and its neighbour ${ q }_{ i }$ have the same semantic label, then ${c}_{p} \oplus { c }_{ q_i }=0$, otherwise ${c}_{p} \oplus { c }_{ q_i }=1$.

As can be seen in Fig~\ref{fig:VoxelsPosition}, voxels that differ significantly from surroundings
get a higher $M_{LGA}$ value than voxels consistent with surrounding voxels.
For instance,
in Fig~\ref{fig:VoxelsPosition}(a), voxels which are inside an object are consistent with the surrounding voxels and have LGA value equal to $0$.
In Fig~\ref{fig:VoxelsPosition}(b), voxels which are at the surface of an object are consistent with the interior voxels but different from other voxels regarding to the semantic categories and have LGA value equal to $1$. 
% LGA value of all the voxels excluding those within the empty space are calculated.
We calculate the LGA value of all the voxels excluding those within the empty space.

The voxels inside the object account for a very high proportion of the overall voxels.
However, the dominant homogeneous voxels with consistent semantic properties are easier to be trained and be classified correctly at a very early stage, and contribute little to the gradient update for the whole training process, which leads to slow convergence of the network.
A histogram of LGA value is shown in Fig~\ref{fig:hist}, where 84.4\% of the voxels with LGA equal to $0$ are in the interior of the object. 
All the remaining voxels with LGA values equal to $1$ to $6$ add up to only 15.6\%.

\subsubsection{PA-Loss}
LGA is used to measure the spatial importance of different voxels and  the position importance aware factor to construct the PA-Loss.
Through this way, the proposed loss function has a strong response to the voxels with rich detail.

Based on LGA values, we set a base value $\lambda$  and a constant $\alpha$ to control LGA importance factor $I$ which is denoted as follows:

\begin{equation}\label{Eq:importancefactor}
{ I }=\lambda  + \alpha { M }_{ LGA }
\end{equation}

PA-Loss is defined as follows:
\begin{equation}\label{Eq:surfaceLoss}  %\Eta
{ L }_{ PA }=-\frac { 1 }{ N } \sum _{ n=1 }^{ N }{ \sum _{ c=1 }^{ C }
{  { I }_{ n }  { y }_{ nc }\log { { \hat { y }  }_{ nc } }  }  } 
\end{equation}
where $N$ is the number of total voxels used to calculate the loss, and $C$ is the number of classes. $ \mathbf{{ y }_{ nc } }$, $ \mathbf{{ \hat { y }  }_{ nc } }$ 
are the one-hot vector of the ground truth labels and the corresponding predictions in class $c$, ${ I }_{ n }$ is the LGA importance factor of the voxel $n$. 

The PA-Loss is differentiable, since the gradient of the PA-Loss can be easily computed from Eq.~\ref{Eq:lga},~\ref{Eq:importancefactor} and Eq.~\ref{Eq:surfaceLoss} which are all differentiable.

%%%%%%%%%%%%%%%%%%%%%%%%%%%%%%%%%%%%%%%%%%%%%%%%%%%%%%%%%%%%%%%%%%%%%%%%%%%%%%%%
\subsection{2D and 3D Hybrid Network Architecture}
The proposed PALNet is composed of four parts as illustrated in Fig~\ref{fig:NetworkStructure}. 
The depth stream~(\ref{section-2Dstream}) takes a full resolution depth as input with a 2D-3D projection layer~(\ref{section-2D3Dprojection}) to transfer the feature tensors in 2D space into 3D space.
The voxel stream~(\ref{section-3Dstream}) employs 3D CNN and takes the 3D grid as input. 
Then, both features extracted from the depth stream and voxel stream are fed into the multi-level feature aggregation module~(\ref{section-Aggregation}).
After that, the reconstruction part~(\ref{section-Reconstruction}) predicts the dense volume to perform the semantic scene completion.

\input{tabs/NYU.tex}

\input{tabs/NYUCAD.tex}
%%%%%%%%%%%%%%%%%%%%%%%%%%%%%%%%%%%%%%%%%%%%%%%%%%%%%%%%%%%%%%%%%%%%%%%%%%%%%%%%
\subsubsection{Depth-Stream}\label{section-2Dstream}

As depicted in the upper left of Fig.~\ref{fig:NetworkStructure}, the depth-stream contains both 2D CNN and 3D CNN modules.
The full resolution depth is first passed a 2D CNN module to extract the 2D features. The 2D CNN module consists two 2D dilated residual blocks with bottle-neck structure (Resblock in Fig.~\ref{fig:NetworkStructure}).
The 3D CNN module consists of a 3D convolution layer, two 3D Resblocks and one 3D pooling layer. 
Through the 2D-3D projection layer, 2D features extracted from depth can be mapped into corresponding 3D space, which helps the network to preserve detailed information.
Within the depth stream, each convolutional layer employs the Residual blocks~\cite{he2016ResNet}.
We adopt the dilated convolution~\cite{yu2015multi} to increase the receptive field of the network.
We also take the advantage of its bottleneck version to increase the capacity of the network and reduce its parameters.
Compared to the pure 3D CNN based approaches, a part of our hybrid network uses 2D CNN, which introduces a small number of parameters but can effectively improve the network depth and capacity.

\subsubsection{2D-3D Projection}\label{section-2D3Dprojection}
% According to the given camera parameters, 
Each pixel in depth can be projected to a voxel in 3D space. 
The mapping index ${\mathcal{T}}_{u,v}$ of a pixel in depth at $(u,v)$ can be computed using the depth value $I_{u,v}$ and the camera parameters which are provided along with each image.
The 2D feature map can be mapped to the 3D space according to the correspondence between the depth and 3D voxels.

% \noindent
\subsubsection{Voxel-Stream}\label{section-3Dstream}
In the voxel-stream, flipped-TSDF~(f-TSDF) ~\cite{song2017_SSCNet, Garbade2018_twoStream} is adopted as the network input.
As depicted in the bottom left of Fig.~\ref{fig:NetworkStructure}, the 3D CNN module contains a 3D convolution, two 3D Resblocks, and one 3D pooling layer. 
Meanwhile, we voxelize the full 3D scene with object labels as the ground truth and the rest voxels in the scene are labeled as empty space resulting in a fully labeled voxel grid representation of the scene.

\subsubsection{Multi-level Feature Aggregation Module}\label{section-Aggregation}
The multi-level feature aggregation module combines the two stream features through element-wise addition. 
Then, the combined features are handled by a series of 3D dilated residual blocks to increase the receptive field and different dilation rates are employed within each block to reduce the gridding problem~\cite{wang2018understanding}.

\subsubsection{Dense Grid Reconstruction}\label{section-Reconstruction}
Finally, in the dense grid reconstruction, there are three standard 3D convolutional layers piled up to give the outputs. 
PALNet predicts the probability distribution of voxel occupancy and object categories for all voxels inside the camera view frustum.

% \subsubsection{PALNet vs SSCNet}
% Compared to SSCNet~\cite{song2017_SSCNet}, PALNet
% (1) adopt bottle-neck residual block (in both feature extractor and multi-level feature aggregation module ) instead of the plan 3D convolution. (2) Part of the hybrid network uses 2D CNN, which introduces a small number of parameters but can effectively improve the network depth and capacity.
% (3) The bottle-neck structure increases the network depth as well.
% (4) decrease the convolution blocks from 3 to 2 within the 3D feature extractor.
% Which together lead to a network with less parameters, but better performance. 

\input{figs/vizResults.tex}

%%%%%%%%%%%%%%%%%%%%%%%%%%%%%%%%%%%%%%%%%%%%%%%%%%%%%%%%%%%%%%%%%%%%%%%%%%%%%%%%

\section{Experimental results}
\label{experiments}

\subsection{Datasets}
We evaluate the proposed method on the NYU~\cite{silberman2012NYUdataset} and NYUCAD~\cite{firman2016NYUCAD} datasets.
The NYU dataset includes 1449 depth maps (795 for training, 654 for testing) captured by the Kinect depth sensor.
The ground truth for completion and semantic segmentation are obtained from~\cite{guo2015predicting} by voxelizing the 3D mesh annotations. 
In some cases, the manually labeled volumes and their corresponding depth are not well aligned in the NYU dataset. 
To address the misalignment, Firman \etal~\cite{firman2016NYUCAD} provide the NYUCAD dataset, in which the depth is rendered from the labeled volume.

%%%%%%%%%%%%%%%%%%%%%%%%%%%%%%%%%%%%%%%%%%%%%%%%%%%%%%%%%%%%%%%%%%%%%%%%%%%%%%%%

\subsection{Implementation Details}
We implement the PALNet with PyTorch and train it from scratch.   
% The dimensions of the 3D space we consider are $4.8$\,m horizontally, $2.88$\,m vertically, and $4.8$\,m in depth. 
The dimensions of the 3D space $\left( H\times D\times V \right) $ are $4.8\times 2.88\times 4.8 $\,m.
We encode the 3D scene into a flipped TSDF~(f-TSDF) with grid size of $0.02$\,m and a truncation value of $0.24$\,m resulting in a volume of resolution $240\times 144\times 240$ as the input of the 3D branch.
For the importance factor $I$, the base value $\lambda$, and the constant $\alpha$ are set to $1.0$ and $0.5$ respectively.

The details of the network architecture are shown in Fig.~\ref{fig:NetworkStructure}.
% including kernel size, dilation rate, output channel, stride used in each convolutional layer. 
During training, we minimize the PA-Loss using SGD with a momentum 0.9 and a weight decay 0.0001. The model is trained with 4 GTX 1080Ti GPUs for 40 epochs with batch size 4. The initial learning rate is set to 0.01, and it is reduced by a factor of 0.1 every 10 epochs.
%%%%%%%%%%%%%%%%%%%%%%%%%%%%%%%%%%%%%%%%%%%%%%%%%%%%%%%%%%%%%%%%%%%%%%%%%%%%%%%%

\subsection{Evaluation Metric}
For semantic scene completion (SSC), we measure the intersection over union (IoU) between the ground truth and the predicted volume, excluding voxels outside the view or the room. The IoU score is calculated for each category and then it is averaged over all the classes to obtain mean IoU (mIoU) score. For the scene completion (SC) task, we treat all non-empty object classes as one category and evaluate precision, recall, and IoU of the binary predictions on occupied voxels. 

\subsection{Comparisons with the State-of-the-arts}
We set the new state-of-the-art performance for both scene completion (SC) and semantic scene completion (SSC) on both NYU~\cite{silberman2012NYUdataset} and NYUCAD~\cite{firman2016NYUCAD} datasets.

% In Table~\ref{tab:results_on_nyu}, we compare PALNet with approaches from Lin \etal~\cite{lin2013holistic}, Geiger and Wang~\cite{geiger2015joint}, SSCNet~\cite{song2017_SSCNet}, TS3D~\cite{Garbade2018_twoStream}, EsscNet~\cite{zhang2018efficient}, and VVNet~\cite{guo2018_VVNet} on NYU dataset, 
We compare PALNet with previous state-of-the-art methods as represented in Table~\ref{tab:results_on_nyu} on NYU dataset and PALNet achieves the best performance in both SC and SSC tasks.
SSCNet and VVNet are pre-trained on the synthetic dataset SUNCG~\cite{song2017_SSCNet}.
% while our method is not. 
Compared to SSCNet, the IoUs of PALNet for SC and SSC tasks increased by 4.7\% and 3.6\%, respectively.
% PALNet surpasses the latest approach VVNet~\cite{guo2018_VVNet} by o.2\% in SC task and  1.2\% in SSC task.
Compared to VVNet~\cite{guo2018_VVNet}, PALNet leads the SC task by 0.2\% and the SSC task by 1.2\%.
% Compared to the methods, DDR-SSC~\cite{li2019rgbd} and TS3D~\cite{Garbade2018_twoStream}, that use additional color information, our method achieves better performance, that is 3.7\% higher and 1.5\% higher in SSC, respectively.
DDR-SSC~\cite{li2019rgbd} and TS3D~\cite{Garbade2018_twoStream} use additional color information.
Our method achieves 3.7\% and 1.5\% higher IoU in SSC than these methods, respectively.
Table~\ref{tab:results_on_nyu} also lists the IoU for each object class, and our method is relatively more accurate than most of the state-of-the-art methods in each category.

% On the NYUCAD dataset, we add two more comparison methods from Zheng \etal~\cite{zheng2013beyond} and Firman \etal~\cite{firman2016NYUCAD}.
We also compare our method with Zheng \etal~\cite{zheng2013beyond} and Firman \etal~\cite{firman2016NYUCAD} on the NYUCAD dataset.
The results are shown in Table~\ref{tab:results_on_nyucad}, and our PALNet achieves the best performance among all the methods as well.

%%%%%%%%%%%%%%%%%%%%%%%%%%%%%%%%%%%%%%%%%%%%%%%%%%%%%%%%%%%%%%%%%%%%%%%%%%%%%%%%

\subsection{Qualitative Results}
Fig.~\ref{fig:vizResults} illustrates the results of SSC task on NYUCAD dataset generated by different methods. 
In Fig.~\ref{fig:vizResults}, (a) is the color image, (b) is the input depth, (c) is the ground truth, (d) and (e) are the results of PALNet and SSCNet~\cite{song2017_SSCNet}. As can be seen, PALNet gives more accurate predictions than SSCNet~\cite{song2017_SSCNet} such as the floor and the chairs in the first row. 
Although the windows in the second row are hard to distinguish, our method still achieves better semantic segmentation results. 
The paintings in the third row also demonstrate the superiority of our approach.
% In the fourth row, due to the lack of texture information, neither SSCNet nor PALNet can correctly identify the furniture beside the chairs. 
In the fourth row, the ground-truth \textit{furniture} circled by the red dashed rectangle is very similar to the object in terms of shape and neither SSCNet nor PALNet can correctly identify the furniture beside the chairs.

\input{tabs/TwoBranches.tex}

\section{Ablation study}
\label{ablationstudy}
We validate the proposed PALNet with a set of ablation tests on NYUCAD~\cite{firman2016NYUCAD} dataset. 
Specifically, we examine three things as listed below:
1) effectiveness of PA-Loss; 
2) hybrid structure \ie combined 2D depth and 3D grid.

%%%%%%%%%%%%%%%%%%%%%%%%%%%%%%%%%%%%%%%%%%%%%%%%%%%%%%%%%%%%%%%%%%%%%%%%%%%%%%%%

% \input{figs/curveSLCompletion.tex}

% \input{figs/curveSLSemantic.tex}

\input{figs/curve.tex}

%%%%%%%%%%%%%%%%%%%%%%%%%%%%%%%%%%%%%%%%%%%%%%%%%%%%%%%%%%%%%%%%%%%%%%%%%%%%%%%%
\subsection{PA-Loss} 
In order to prove the effectiveness of the proposed Position Aware Loss(PA-Loss), we design two sets of comparative experiments. 
Firstly, we train PALNet with different loss functions (PA-Loss and other loss functions) to compare their training results. 
Then, we apply PA-Loss to SSCNet to demonstrate that PA-Loss can improve the accuracy of SSCNet. 
The details of these ablation studies are presented as follows.

\subsubsection{Comparison with Different Loss Functions} 
The three comparison methods are weighted cross-entropy (WCE) loss, focal loss and dice loss.

\noindent
\textbf{Weighted Cross-entropy Loss.} 
The weighted cross-entropy loss is defined as follows:
\begin{equation}\label{Eq:weightedcrossEntropyLoss} 
{ L }_{ WCE }=-\frac { 1 }{ N } \sum _{ n=1 }^{ N }{ \sum _{ c=1 }^{ C }{ { w }_{ c } { y }_{ nc }\log { { \hat { y }  }_{ nc } }  }  } 
\end{equation}
where $N$ is the number of total voxels, $C$ is the number of classes. 
Voxels belong to each class have different weights ${ w }_{ c }$. Specifically, the ratio of each category is counted including empty voxels on the training set and use the reciprocal of the ratio is used as the weight of each category. ${ \hat {y} }_{ nc }$ and ${ y }_{ nc }$ denote the prediction and ground truth of voxel $n$ belong to class $c$.
% The weighted summation runs over the predicted segmentation volume ${ \hat {y} }_{ nc }$ and the ground truth volume ${ y }_{ nc }$.

\noindent
\textbf{Focal Loss.} 
Focal loss adds a factor $(1-p_t)^\gamma$ to the standard cross-entropy criterion, as follows:
\begin{equation}\label{Eq:focalloss}  
{ L }_{ FL }=-\frac { 1 }{ N } \sum _{ n=1 }^{ N }{ \sum _{ c=1 }^{ C }{ { \left( 1-{ \hat { y }  }_{ nc } \right)  }^{ \gamma  }{ y }_{ nc }\log { { \hat { y }  }_{ nc } }  }  } 
\end{equation}
In the experiment, we set $\gamma = 2$ as it is suggested to reduce the relative weights for easy classified examples.

\noindent
\textbf{Multi-class Dice Loss.} % 
% Standard dice loss can only be used for the binary classification. 
We generalize the standard dice loss for multi-class segmentation by applying binary version on each class iteratively. 
The multi-class dice loss is defined as:
\begin{equation}\label{Eq:multidiceloss} 
L_{ MDL }=\sum _{ c=1 }^{ C }{ 1-\frac { 2\sum _{ n=1 }^{ N } y_{ nc }\hat { y } _{ nc } }{ \sum _{ n=1 }^{ N } y_{ nc }^{ 2 }+\sum _{ n=1 }^{ N } \hat { y } _{ nc }^{ 2 } }  } 
\end{equation}

\input{tabs/AS_Accuracy_Different_Loss-Ours.tex}

We use these three loss functions to train the PALNet from scratch and evaluate the accuracy of the network after each epoch.
As shown in Table.~\ref{tab:PALNet_with_4_loss}, the network trained with PA-Loss provides more accurate results than that trained with other loss functions on both SC and SSC tasks.
% As shown in Table.~\ref{tab:PALNet_with_4_loss}, all the networks that used these four loss functions for training have achieved very good results. Among them, the network using PA-Loss for training is the best. 
In the task of semantic scene completion, PALNet trained with PA-Loss achieves 0.3\%, 1.3\% and 0.5\% higher mIoU than that trained with weighted cross-entropy loss, dice loss and focal loss.

In addition, we find that training convergence speed of using PA-Loss is slightly faster than that of other loss functions.
We plot the testing accuracy of each epoch as a curve, as shown in Fig.~\ref{fig:curve} for scene completion (left) and semantic scene completion (right).

\input{tabs/AS_Accuracy_Different_Loss-SSCNet.tex}

\subsubsection{Apply PA-Loss to SSCNet} 

To further verify the effectiveness and generalization ability of PA-Loss, we apply PA-Loss to SSCNet~\cite{song2017_SSCNet} in SSC task.
We reimplemented SSCNet in PyTorch and denote our implementation as SSCNet* in the following discussions.
We train
SSCNet* with WCE Loss and PA-Loss separately and evaluate the performance of SSCNet* on NYUCAD dataset. 
The experimental results are shown in Table.~\ref{tab:loss-SSCNet}. The results reported in SSCNet~\cite{song2017_SSCNet} are also listed for reference. Our reimplemented SSCNet* gets comparable (slightly better) accuracy than SSCNet~\cite{song2017_SSCNet}.
Compared to SSCNet* trained with WCE Loss, the one trained with our PA-Loss gets better accuracy on both SC and SSC tasks.

\subsection{Hybrid Structure Combined 2D Depth and 3D Voxels}
We explore the contribution of the 2D and 3D parts of the hybrid structure.
To investigate the effect of each stream, we develop two variants of the PALNet where each variant contains only one of the two input branches.
Note that the element-wise add operation is no longer needed in the single-stream network and the feature maps are fed into the multi-level feature aggregation stage immediately.
Since the 3D branch is extended from SSCNet
% by replacing its residual block with the bottle-neck version
, therefore, to conclusively exemplify our method we select SSCNet for comparison in this section.

\input{tabs/Params.tex} 
As shown in Table~\ref{tab:results_2barnch}, both variants with one stream as input achieve better performance than SSCNet~\cite{song2017_SSCNet} on the SC and SSC tasks. The combination of 2D and 3D streams increases the accuracy of PALNet in SC to 80.8\% and SSC to 46.6\%.

The variant network with 2D depth as the input achieves 46.1\% IoU on the SSC task and is 2.6\% higher than that of the other variant with 3D voxels as the input.
The significant advantages of the 2D network over 3D network strongly suggest that the details in depth are very useful for semantic scene completion.
The 3D branch with manually designed f-TSDF gets better result (0.3\%) than the 2D branch on the task of scene completion. This is reasonable due to the carefully designed f-TSDF encodes the geometry information well. 
The symbolic information of f-TSDF can explicitly tell the network where the occluded space is, and help the network to pay attention to these parts for corresponding shape completion.

In Table~\ref{tab:SpeedMemory}, the parameters, Flops, inference speed, memory and accuracy of the networks among SSCNet~\cite{song2017_SSCNet}
% , DDR-SSC~\cite{li2019rgbd} 
and PALNet are listed out. 
By taking advantage of the bottle-neck residual network, PALNet has much deeper structure than SSCNet~\cite{song2017_SSCNet}
% , and achieves the best trade-off compared with other two baseline methods. 
and achieve better accuracy.
%%%%%%%%%%%%%%%%%%%%%%%%%%%%%%%%%%%%%%%%%%%%%%%%%%%%%%%%%%%%%%%%%%%%%%%%%%%%%%%%
\section{Conclusions}
\label{conclusion}
We introduce PALNet which takes both the depth and TSDF as inputs for semantic scene completion. 
The feature from 2D-stream are projected and concatenated with the feature in 3D-stream and trained in an end-to-end manner. 
We also propose a position importance aware loss, PA-Loss, which leads to slightly faster training convergence and better performance.
The experiments on both synthetic and real datasets validate the superior performance of the proposed method. 
The two interesting topics that are worth exploring in our future work include: (1) a better trade-off between 2D CNNs and 3D CNNs to employ less 3D convolutional layers without sacrificing the performance, and (2) to extend our PALNet framework with RGB-D as input.
% 
%%%%%%%%%%%%%%%%%%%%%%%%%%%%%%%%%%%%%%%%%%%%%%%%%%%%%%%%%%%%%%%%%%%%%%%%%%%%%%%%

\section*{Acknowledgement}
We gratefully acknowledge the support of the Australian Research Council through the Centre of Excellence for Robotic Vision CE140100016 and Laureate Fellowship FL130100102 to IR. This work is also supported by the National Science Foundation of China under Grants 61773210 and 61603184.

%%%%%%%%%%%%%%%%%%%%%%%%%%%%%%%%%%%%%%%%%%%%%%%%%%%%%%%%%%%%%%%%%%%%%%%%%%%%%%%%

% \addtolength{\textheight}{-12cm}  

%%%%%%%%%%%%%%%%%%%%%%%%%%%%%%%%%%%%%%%%%%%%%%%%%%%%%%%%%%%%%%%%%%%%%%%%%%
% \section*{APPENDIX}

% Appendixes should appear before the acknowledgment.

%%%%%%%%%%%%%%%%%%%%%%%%%%%%%%%%%%%%%%%%%%%%%%%%%%%%%%%%%%%%%%%%%%%%%%%%%%%%%%%%

\bibliographystyle{IEEEtran}

%\bibliography{IEEEabrv,SscReference}
\bibliography{IEEEabrv, root}

\end{document}

%% file: figs/teaser.tex
\begin{figure}[htbp]
\centering{\includegraphics[width=0.95\columnwidth]{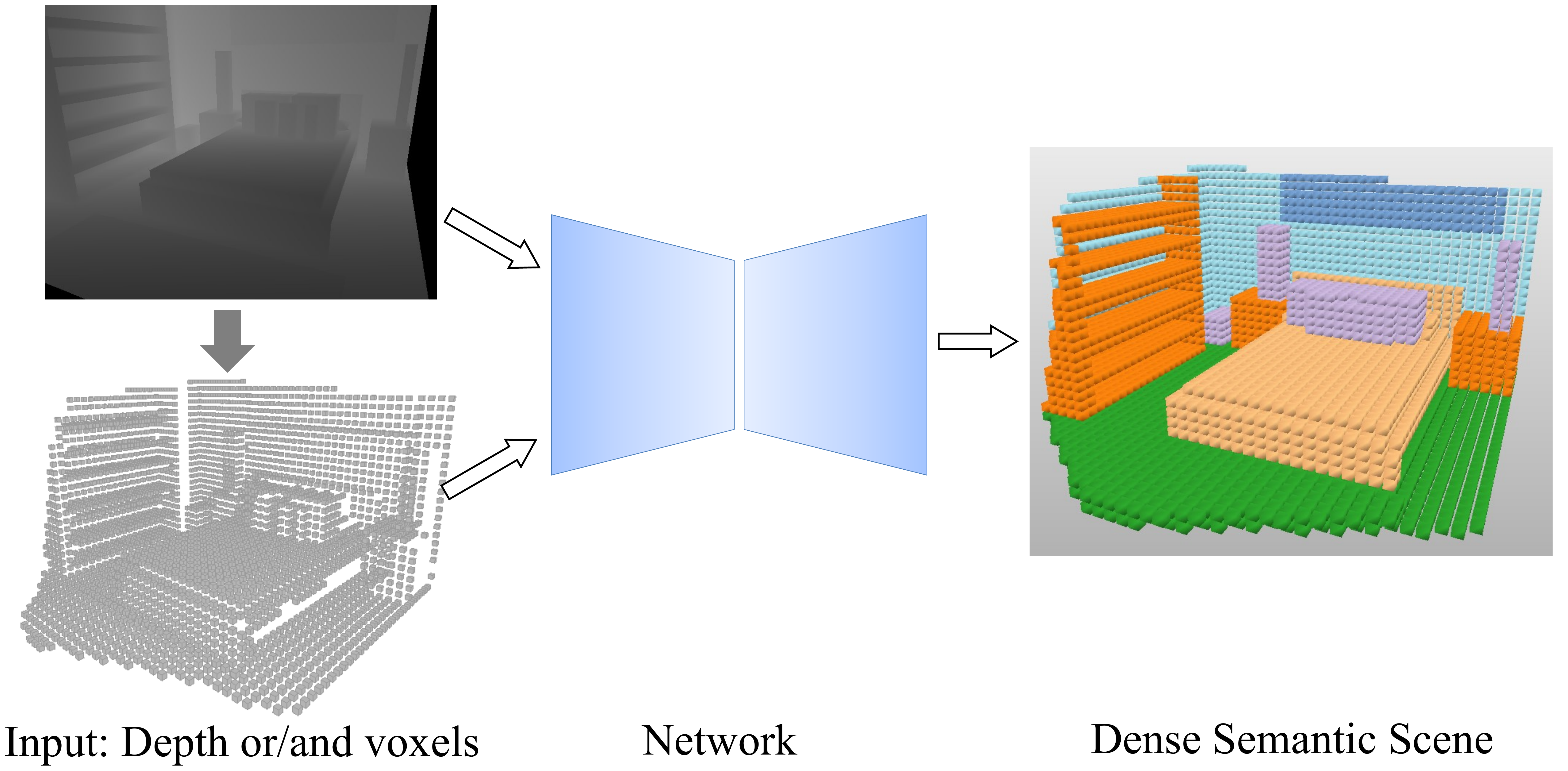}}
% \vspace{-2mm}
\caption{Semantic Scene Completion: Based on input depth or its corresponding 3D voxel volume, the goal of SSC is to simultaneously complete the partial 3D shapes and predict the dense semantic labels of both observed and unobserved parts in the view frustum.} 
% \vspace{-5mm}
\label{fig:teaser}
\end{figure}

%% file: figs/VoxelPosition.tex
% 
% 
\begin{figure*}[ht]
\centering
{
\includegraphics[width=0.95\linewidth]{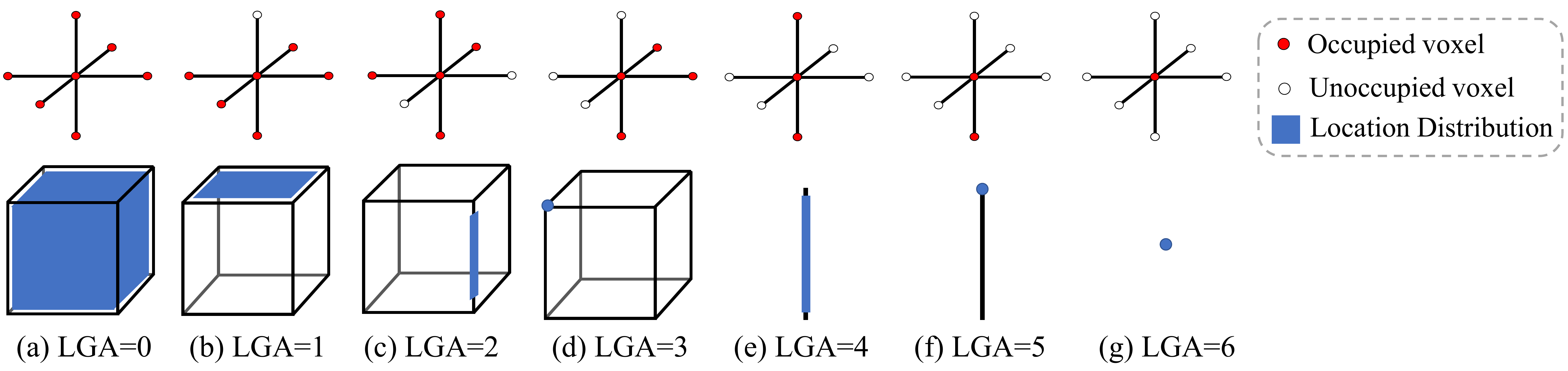}
}
% \vspace{-0.3cm}
\caption
{Local Geometric Anisotropy: voxels at different positions contain different local geometric information.
(a)-(d) respectively indicate voxels inside the object, voxels on the surface, voxels on the edge, and a voxel on the vertex. For a strip-shaped object, (e),(f) indicate voxels on the edge and a voxel on the vertex.
(g) indicates an isolated voxel.
}
\label{fig:VoxelsPosition}
\end{figure*}

%% file: figs/hist_percentage.tex
\begin{figure}[ht]
\centering
{
\includegraphics[width=\columnwidth]{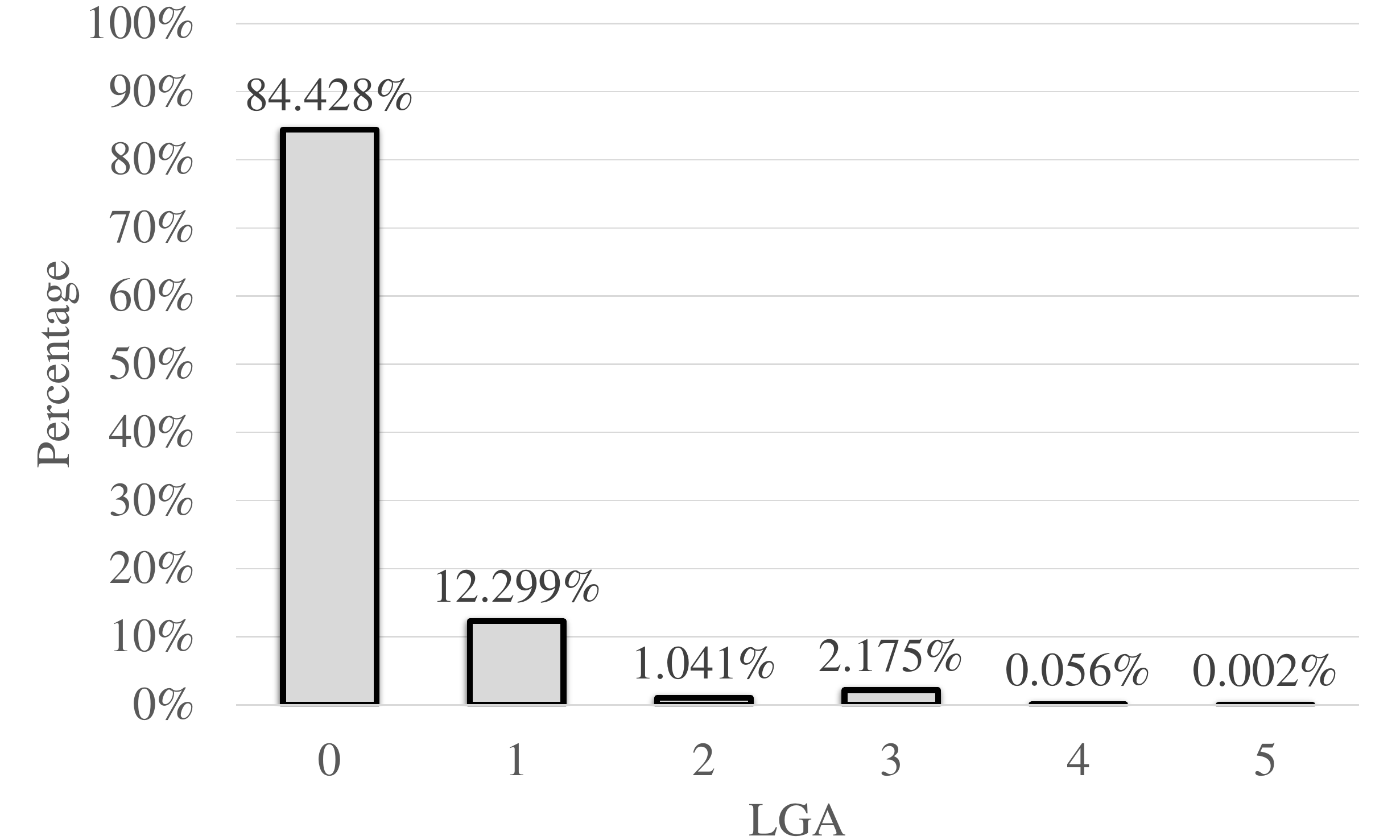}
}
% \vspace{-8mm}
\caption{
The percentage of voxels with different LGA values in dataset NYU~\cite{silberman2012NYUdataset}. 
More than 84\% of the voxels have $LGA=0$, indicating that most of the voxels are inside the objects, and less than 16\% of the voxels are outside of the objects. Voxels in free space are not taken into account.
}\label{fig:hist}
\end{figure}

%% file: figs/NetworkStructure.tex
\begin{figure*}[ht]
\centering
{
\includegraphics[width=\linewidth]{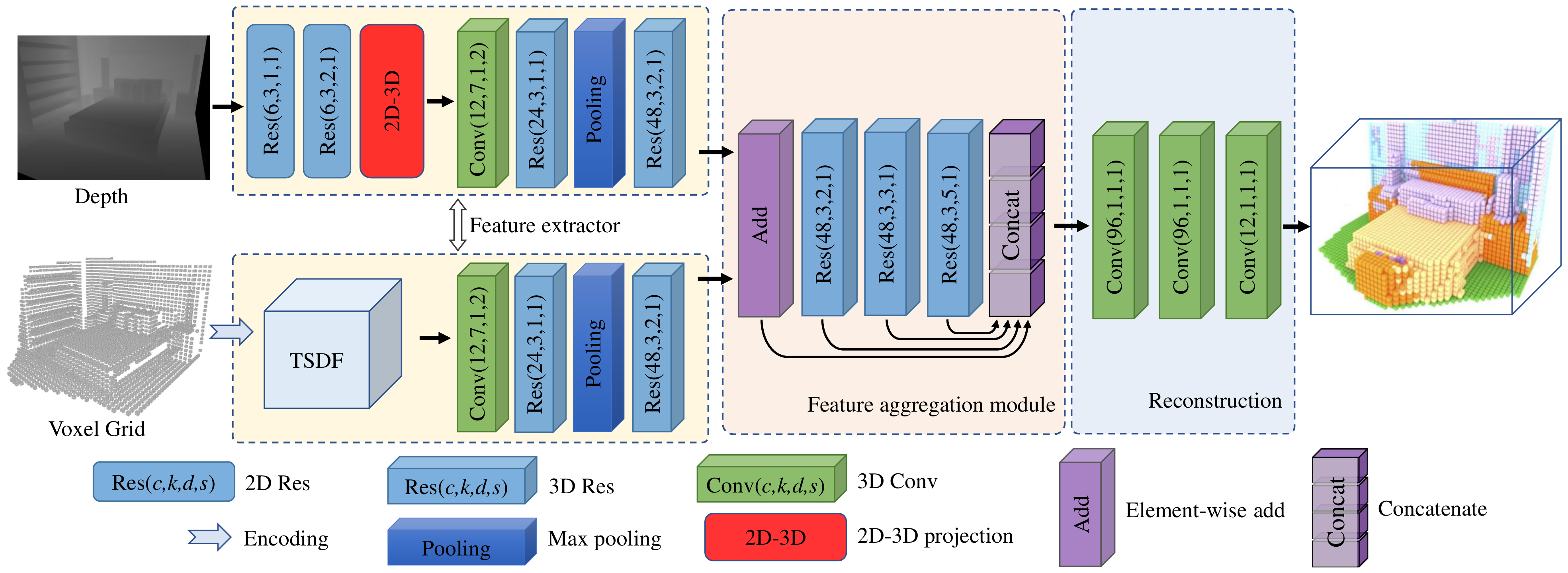}
}
% \vspace{-0.7cm}
\caption{The hybrid network structure of PALNet for semantic scene completion. In depth-stream, 2D CNN is used to process the full resolution depth, the feature maps are projected to 3D space with the 2D-3D projection layer, and followed by 3D convolutions. In voxel-stream, TSDF encoded voxels are sent as input, all convolution operations employed are 3D convolutions. After the two streams of information are aggregated, they are sent to the subsequent network with large receptive field to capture 3D context to predict the complete semantic scene.
We represent the 3D operation in the form of a cuboid in this figure.
Res($c,k,d,s$) is the dilated residual blocks with bottleneck. $c$ is the output channel, $k$ is the kernel size, $d$ is the dilation rate, and $s$ is stride in the convolution. Conv($c,k,d,s$) is the 3D convolutional layer.}
\label{fig:NetworkStructure}
\end{figure*}

%% file: tabs/NYU.tex
\begin{table*}[ht]
\caption{Results of various methods on NYU dataset~\cite{silberman2012NYUdataset}. Bold numbers represent the best score.}
% \vspace{-1mm}
\centering
{\begin{tabular}
{l |c c c|c c c c c c c c c c c c} 
\hline
& \multicolumn{3}{c|}{scene completion} & \multicolumn{12}{c}{semantic scene completion} \\ \hline
Method  & prec. & recall & IoU & \cellcolor{rgb1}ceil. & \cellcolor{rgb2}floor & \cellcolor{rgb3}wall & \cellcolor{rgb4}win. & \cellcolor{rgb5}chair & \cellcolor{rgb6}bed & \cellcolor{rgb7}sofa & \cellcolor{rgb8}table & \cellcolor{rgb9}tvs & \cellcolor{rgb10}furn. & \cellcolor{rgb11}objs. & mIoU \\ 
\hline
Lin et al.~\cite{lin2013holistic}   & 58.5 & 49.9 & 36.4 &  0.0 & 11.7 & 13.3 & {\bfseries 14.1} &  9.4 & 29.0 & 24.0 &  6.0 &  7.0 & 16.2 &  1.1 & 12.0\\ %& 2013NYU
Geiger et al.~\cite{geiger2015joint} & 65.7 & 58.0 & 44.4 & 10.2 & 62.5 & 19.1 &  5.8 &  8.5 & 40.6 & 27.7 &  7.0 &  6.0 & 22.6 &  5.9 & 19.6\\ %&2015NYU
    \hline
SSCnet~\cite{song2017_SSCNet} & 59.3 & {\bfseries 92.9} & 56.6 & 15.1 & 94.6 & 24.7 & 10.8 & 17.3 & 53.2 & 45.9 & 15.9 & 13.9 & 31.1 & 12.6 & 30.5\\ %2017
TS3D~\cite{Garbade2018_twoStream}     & 64.9 & 88.8 & 60.2 & 8.2 & 94.1 & 26.4 & 19.2 & 17.2 & 55.5 & 48.4 & 16.4 & 22.0 & 34.0 & {\bfseries 17.1} & 32.6\\   % v2
EsscNet~\cite{zhang2018efficient}   & {\bfseries 71.9} & 71.9 & 56.2 & 17.5 & 75.4 & 25.8 &  6.7 & 15.3 & 53.8 & 42.4 & 11.2 &    0 & 33.4 & 11.8 & 26.7\\ 
DDR-SSC~\cite{li2019rgbd}  & 71.5  & 80.8 & 61.0 & 21.1 & 92.2 & {\bfseries 33.5} & 6.8 & 14.8 & 48.3 & 42.3 & 13.2 &  13.9 & 35.3 & 13.2 & 30.4 \\ 
VVNet~\cite{guo2018_VVNet}  & 69.8 & 83.1 & 61.1 & 19.3 & {\bfseries 94.8} & 28.0 & 12.2 & 19.6 & {\bfseries 57.0} & {\bfseries 50.5} & {\bfseries 17.6} & 11.9 & 35.6 & 15.3 & 32.9\\ 
\hline
PALNet(ours)      &  68.7 & 85.0 & {\bfseries 61.3} & {\bfseries 23.5} & 92.0 & 33.0 & 11.6 & {\bfseries 20.1} & 53.9 & 48.1 & 16.2 & {\bfseries 24.2} & {\bfseries 37.8} & 14.7 & {\bfseries 34.1}\\ 
\hline
\end{tabular}\label{tab:results_on_nyu}}
\end{table*}

%% file: tabs/NYUCAD.tex
\begin{table*}[ht]
\caption{
Results of various methods on NYUCAD dataset~\cite{firman2016NYUCAD}. Bold numbers represent the best score.
}
% \vspace{-1mm}
\centering
\scalebox{1.0}
{\begin{tabular}
{l |c c c|c c c c c c c c c c c c}  
\hline
& \multicolumn{3}{c|}{scene completion} & \multicolumn{12}{c}{semantic scene completion} \\ \hline
Method  & prec. & recall & IoU & \cellcolor{rgb1}ceil. & \cellcolor{rgb2}floor & \cellcolor{rgb3}wall & \cellcolor{rgb4}win. & \cellcolor{rgb5}chair & \cellcolor{rgb6}bed & \cellcolor{rgb7}sofa & \cellcolor{rgb8}table & \cellcolor{rgb9}tvs & \cellcolor{rgb10}furn. & \cellcolor{rgb11}objs. & mIoU \\ 
\hline
Zheng et al.~\cite{zheng2013beyond}    & 60.1 & 46.7 & 34.6 & - & - & - & - & - & - & - & - & - & - & - & - \\
Firman et al.~\cite{firman2016NYUCAD}  & 66.5 & 69.7 & 50.8 & - & - & - & - & - & - & - & - & - & - & - & - \\
\hline
%   SSCnet        & 75.0 & 92.3 & 70.3 & - \\   % trained on NYUCAD
SSCnet~\cite{song2017_SSCNet}    & 75.4 & {\bfseries96.3} & 73.2  & 32.5 & 92.6 & 49.2 &  8.9 & 33.9 & 57.0 & 59.5 & 28.3 &  8.1 & 44.8 & 25.1 & 40.0 \\   % trained on SUNCG and NYUCAD
TS3D~\cite{Garbade2018_twoStream}& 80.2 & 94.4 & 76.5  & 34.4 & {\bfseries93.6} & 47.7 & {\bfseries31.8} & 32.2 & {\bfseries65.2} & 54.2 & 30.7 & {\bfseries32.5} & {\bfseries50.1} & 30.7 & 45.7 \\  
DDR-SSC~\cite{li2019rgbd}   & {\bfseries 88.7} & 88.5 & 79.4 & 54.1 & 91.5 & 56.4 & 14.9 & 37.0 & 55.7 & 51.0 &  28.8 & 9.2 & 44.1 &  27.8 & 42.8 \\
VVNet~\cite{guo2018_VVNet}      & 86.4 & 92.0 & 80.3   & - & - & - & - & - & - & - & - & - & - & - & - \\ 
\hline
PALNet    & 87.2 & 91.7 & {\bfseries80.8}   & {\bfseries54.8} & 92.8 & {\bfseries60.3} & 15.3 & {\bfseries43.1} & 60.7 & {\bfseries59.9} & {\bfseries37.6} &  8.1 & 48.6 & {\bfseries31.7} & {\bfseries46.6} \\
\hline
\end{tabular}\label{tab:results_on_nyucad}}
\end{table*}

%% file: figs/vizResults.tex
%
%
\begin{figure*}[htbp]
\centering{\includegraphics[width=0.8\linewidth]{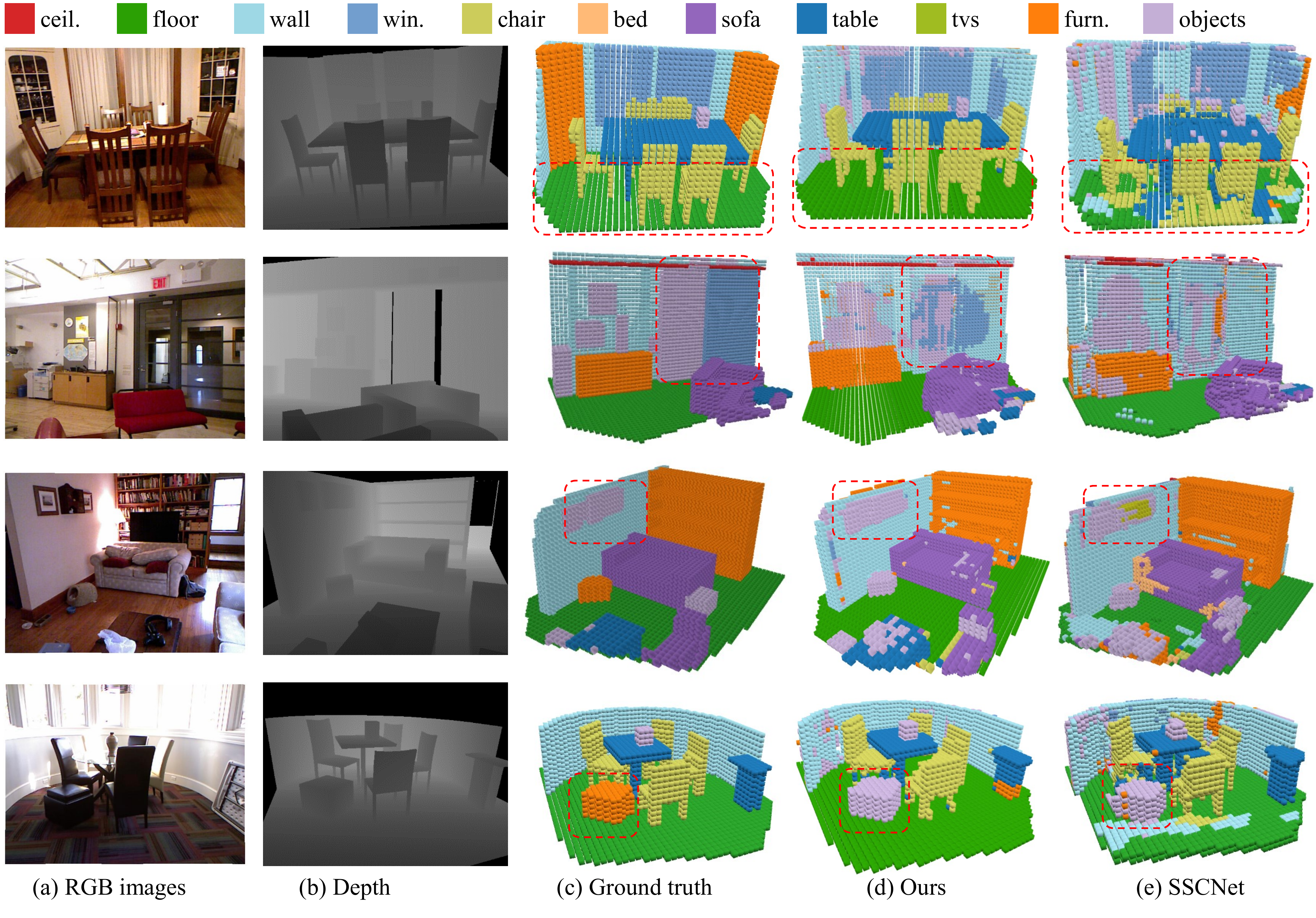}}
% \vspace{-2mm}
\caption{For a scene shown in (a), Semantic Scene Completion (SSC) takes (b) depth image as input and output dense 3D semantic volume. (c) gives the ground truth. (d) and (e) are the predictions of the proposed PALNet and SSCNet~\cite{song2017_SSCNet}.
}
% \vspace{-2mm}
\label{fig:vizResults}
\end{figure*}

%% file: tabs/TwoBranches.tex
\begin{table*}[htbp]
\caption{
% Ablation study for the hybrid structure. Results are tested on the NYUCAD dataset~\cite{firman2016NYUCAD}.
Results of variant PALNets and SSCNet~\cite{song2017_SSCNet}.
}
% \vspace{-2mm}
\centering
\scalebox{1.0}
{\begin{tabular}
{l |c c c|c c c c c c c c c c c c}  
\hline
& \multicolumn{3}{c|}{scene completion} & \multicolumn{12}{c}{semantic scene completion} \\ \hline
Method  & prec. & recall & IoU & \cellcolor{rgb1}ceil. & \cellcolor{rgb2}floor & \cellcolor{rgb3}wall & \cellcolor{rgb4}win. & \cellcolor{rgb5}chair & \cellcolor{rgb6}bed & \cellcolor{rgb7}sofa & \cellcolor{rgb8}table & \cellcolor{rgb9}tvs & \cellcolor{rgb10}furn. & \cellcolor{rgb11}objs. & avg. \\ 
\hline

SSCnet~\cite{song2017_SSCNet}    & 75.4 & {\bfseries96.3} & 73.2  & 32.5 & 92.6 & 49.2 &  8.9 & 33.9 & 57.0 & 59.5 & 28.3 &  8.1 & 44.8 & 25.1 & 40.0 \\   % trained on SUNCG and NYUCAD
PALNet-3D                     & 83.3 & 94.5 & 79.1   & 49.1 & 92.3 & 56.2 &  6.8 & 36.6 & {\bfseries62.5} & 57.6 & 34.6 &  5.9 & 45.6 & 30.7 & 43.5 \\ 
PALNet-2D                     & 83.5 & 93.4 & 78.8   & 52.2 & 92.0 & 56.6 & {\bfseries20.1} & 40.2 & 61.1 & 58.8 & 33.9 & {\bfseries17.1} & 45.3 & 30.0 & 46.1 \\ 
PALNet-hybrid    & {\bfseries87.2} & 91.7 & {\bfseries80.8}   & {\bfseries54.8} & {\bfseries92.8} & {\bfseries60.3} & 15.3 & {\bfseries43.1} & 60.7 & {\bfseries59.9} & {\bfseries37.6} &  8.1 & {\bfseries48.6} & {\bfseries31.7} & {\bfseries46.6} \\
\hline
\end{tabular}\label{tab:results_2barnch}}
% \vspace{-2mm}
\end{table*}

%% file: figs/curve.tex
%
%
%
\begin{figure}[ht]
\centering{\includegraphics[width=1.0\columnwidth]{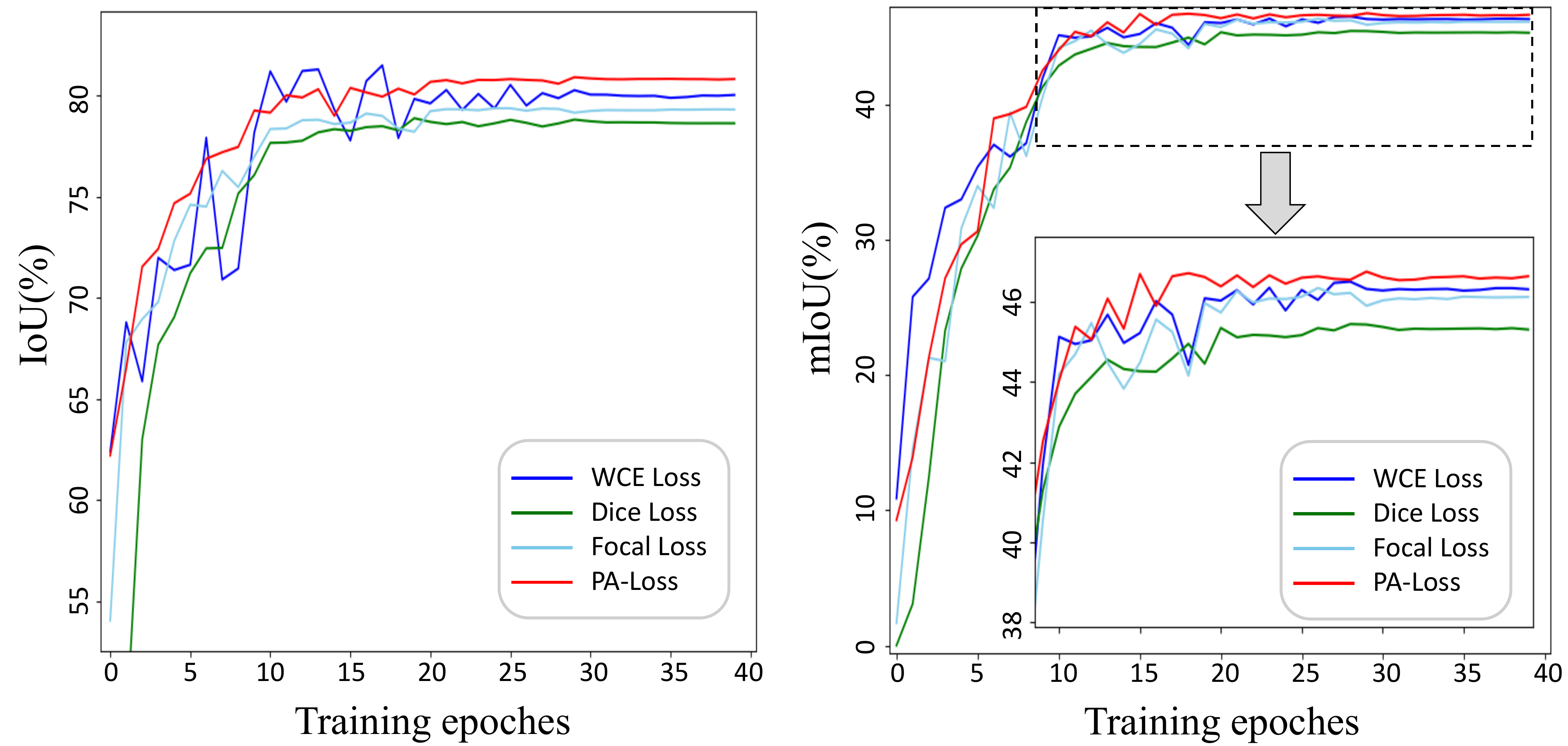}}
% \vspace{-1mm}
\caption{
 Training curves of network with different loss functions for scene completion (left) and semantic scene completion (right).
}
\label{fig:curve}
\end{figure}

%% file: tabs/AS_Accuracy_Different_Loss-Ours.tex
\begin{table}[tb]
\caption{Results of PALNet trained with various loss functions.
}
% \vspace{-1mm}
\centering
\begin{tabular}{l|lll|l}
\hline
Loss function & \multicolumn{1}{c}{Precision} & \multicolumn{1}{c}{Recall} & \multicolumn{1}{c|}{IoU} & \multicolumn{1}{c}{mIoU} \\ \hline
WCE Loss    & 79.4               & {\bfseries 95.9}   & 79.9            & 46.3                  \\
Dice Loss   & {\bfseries 88.2}   & 83.3             & 78.64             & 45.3                  \\
Focal Loss   & 85.6               & 91.8            & 79.3              & 46.1                  \\
PA-Loss   & 87.2                  & 91.7            & {\bfseries 80.8}  & {\bfseries 46.6 }     \\ \hline
\end{tabular}\label{tab:PALNet_with_4_loss}
% \vspace{-2mm}
\end{table}

%% file: tabs/AS_Accuracy_Different_Loss-SSCNet.tex
\begin{table}[tb]
\caption{Results of SSCNet trained with PA-Loss and Weighted Cross-Entropy Loss (WCE Loss). With SSCNet* denotes our implementation of SSCNet~\cite{song2017_SSCNet}.
}
% \vspace{-1mm}
\centering
\scalebox{1.0}
{
\begin{tabular}{l|l|lll|l}
\hline
Methods & Loss function & Precision         & Recall            & IoU               & mIoU              \\ \hline
SSCNet  & WCE Loss          & 75.4              & {\bfseries 96.3}  & 73.2              & 40.0              \\
SSCNet* & WCE Loss          & 76.5              & 95.7              & 74.8              & 42.1              \\
SSCNet* & PA-Loss            & {\bfseries 81.6}  & 91.6              & {\bfseries 76.0} & {\bfseries 43.4}   \\ \hline
\end{tabular}
\label{tab:loss-SSCNet}
}
% \vspace{-2mm}
\end{table}

%% file: tabs/Params.tex
% \vspace{-0.3cm}
\begin{table}[!tb]
\caption{
The number of parameters, inference speed and GPU memory usage of SSCNet~\cite{song2017_SSCNet}
% , DDR-SSC~\cite{li2019rgbd} 
and the proposed PALNet.
}
% \vspace{-2mm}
\begin{center}
\scalebox{0.95}
{
\begin{tabular} {l|c|c|c|c|c|c} 
\hline
\multicolumn{1}{c|}{Method} & \multicolumn{1}{c|}{\begin{tabular}[c]{@{}c@{}}Params\\ (k)\end{tabular}} & \multicolumn{1}{c|}{\begin{tabular}[c]{@{}c@{}}FLOPs\\ (G)\end{tabular}} & \multicolumn{1}{c|}{\begin{tabular}[c]{@{}c@{}}Speed\\ (FPS)\end{tabular}} & \multicolumn{1}{c|}{\begin{tabular}[c]{@{}c@{}}Memory\\ (M)\end{tabular}} & \multicolumn{1}{c|}{\begin{tabular}[c]{@{}c@{}}Network \\ Depth\end{tabular}} &
\multicolumn{1}{c}{\begin{tabular}[c]{@{}c@{}}SSC\\ IoU(\%)\end{tabular}}         \\  \hline
SSCNet 	& 930.0             & 163.8         & 0.7          & 5305            &    14       & 40.0          \\
% DDR-SSC        & \textbf{195.0}    & \textbf{27.2} & \textbf{1.5} & \textbf{1829} & \textbf{44}    & 30.4          \\
% PALNet        & 223.0             & 78.8          & 1.2  & 3717  & 25  & \textbf{46.6}          \\
PALNet        & \textbf{223.0}             & \textbf{78.8}          & \textbf{1.2}  & \textbf{3717}  & \textbf{25}  & \textbf{46.6}          \\
\hline
\end{tabular}
}
\label{tab:SpeedMemory}
\end{center}
% \vspace{-2mm}
\end{table}
% \vspace{-0.6cm}